\documentclass[11pt]{article}

\usepackage[preprint]{acl}

\usepackage{times}
\usepackage{latexsym}

\usepackage[T1]{fontenc}

\usepackage[utf8]{inputenc}

\usepackage{microtype}

\usepackage{inconsolata}

\usepackage{graphicx}

\usepackage{booktabs}
\usepackage{capt-of}
\usepackage{amsmath}
\usepackage{amssymb}
\usepackage{algorithm}
\usepackage{algpseudocode}
\usepackage{multirow}
\usepackage{xcolor}
\usepackage{enumitem}
\usepackage{amsthm}
\newtheorem{definition}{Definition}

\usepackage{tcolorbox}
\tcbuselibrary{breakable,listings,skins}
\definecolor{promptframe}{RGB}{48,65,78}
\definecolor{promptbackground}{RGB}{247,249,250}
\newtcblisting{prompttemplate}[1]{
  enhanced,breakable,listing only,listing engine=listings,
  title={#1},colback=promptbackground,colframe=promptframe,
  colbacktitle=promptframe,coltitle=white,
  fonttitle=\sffamily\bfseries\small,
  sharp corners,boxrule=0.55pt,
  left=3mm,right=3mm,top=2mm,bottom=2mm,
  before skip=8pt,after skip=8pt,
  listing options={basicstyle=\ttfamily\footnotesize,
    breaklines=true,breakatwhitespace=false,columns=fullflexible,
    keepspaces=true,showstringspaces=false,tabsize=2}
}

\newcommand{\CURRENT}{\textsc{Current}}
\newcommand{\STALE}{\textsc{Stale}}
\newcommand{\UNCERTAIN}{\textsc{Uncertain}}
\newcommand{\HISTORICAL}{\textsc{Historical}}
\newcommand{\RECOMPUTE}{\textsc{Needs-Recompute}}
\newcommand{\STRICT}{\textsc{Strict}}
\newcommand{\WEAK}{\textsc{Weak}}
\newcommand{\NODEP}{\textsc{None}}

\title{A State Change Is Local, Its Consequences Are Not:\\
Dependency-Aware State Revision for Long-Term Agent Memory}

\author{
  \mdseries\mbox{%
  Yiqi Wang\textsuperscript{1,*}\enspace
  Tongyuan Zhao\textsuperscript{1}\enspace
  Jiaqi Zhang\textsuperscript{2}\enspace
  Yiqun Duan\textsuperscript{3}\enspace
  Zhangkai Wu\textsuperscript{4}\enspace
  Mingkai Zheng\textsuperscript{5}\enspace
  Taotao Cai\textsuperscript{1}}\\[0.3em]
  \mdseries\small
  \textsuperscript{1}University of Southern Queensland\quad
  \textsuperscript{2}Jiangsu University\quad
  \textsuperscript{3}Uploading Inc.\\
  \mdseries\small
  \textsuperscript{4}The University of Sydney\quad
  \textsuperscript{5}Southern University of Science and Technology\\[0.2em]
  \mdseries\small\textsuperscript{*}\texttt{yiqi.wang.jennie@gmail.com}
}

\begin{document}
\maketitle
\addtocontents{toc}{\protect\setcounter{tocdepth}{-1}}
\raggedbottom

\begin{abstract}
Long-lived agents rely on persistent memory of users, tasks and decisions, and this memory must stay consistent with a world that keeps changing. When an upstream fact changes, however, memory systems can revise the contradicted memory itself while the beliefs, plans and actions derived from it remain accepted, even though they have lost their support. We therefore formulate dependency-aware state revision, which asks not only what changed but also which dependent states should no longer be trusted. To address it, StateGraph maintains an evolving state dependency graph whose edges are recovered from interaction histories. Its core design is counterfactual verification: each candidate edge is tested by asking whether the dependent state would survive the loss of its premise, which separates support dependence from semantic relatedness, and revision then invalidates a dependent state only when a strict dependency fails and no independent support path survives. To test this design, we introduce StateChangeBench, a controlled benchmark of 50 author-constructed cases annotated with cascade depth, hard negatives and action consequences. On it, StateGraph's Impact-F1 margin over a CUPMem reimplementation is 6.9 percentage points at depth 1 and 26.0 at depth 3+. On public long-term memory benchmarks, its point estimates are comparable to or higher than those of the evaluated baselines on most datasets. These results provide initial evidence that explicit dependency tracking can improve selective memory revision under changing conditions.
\end{abstract}

\section{Introduction}
\label{sec:intro}

\begin{figure*}[!t]
  \centering
  \includegraphics[width=\textwidth]{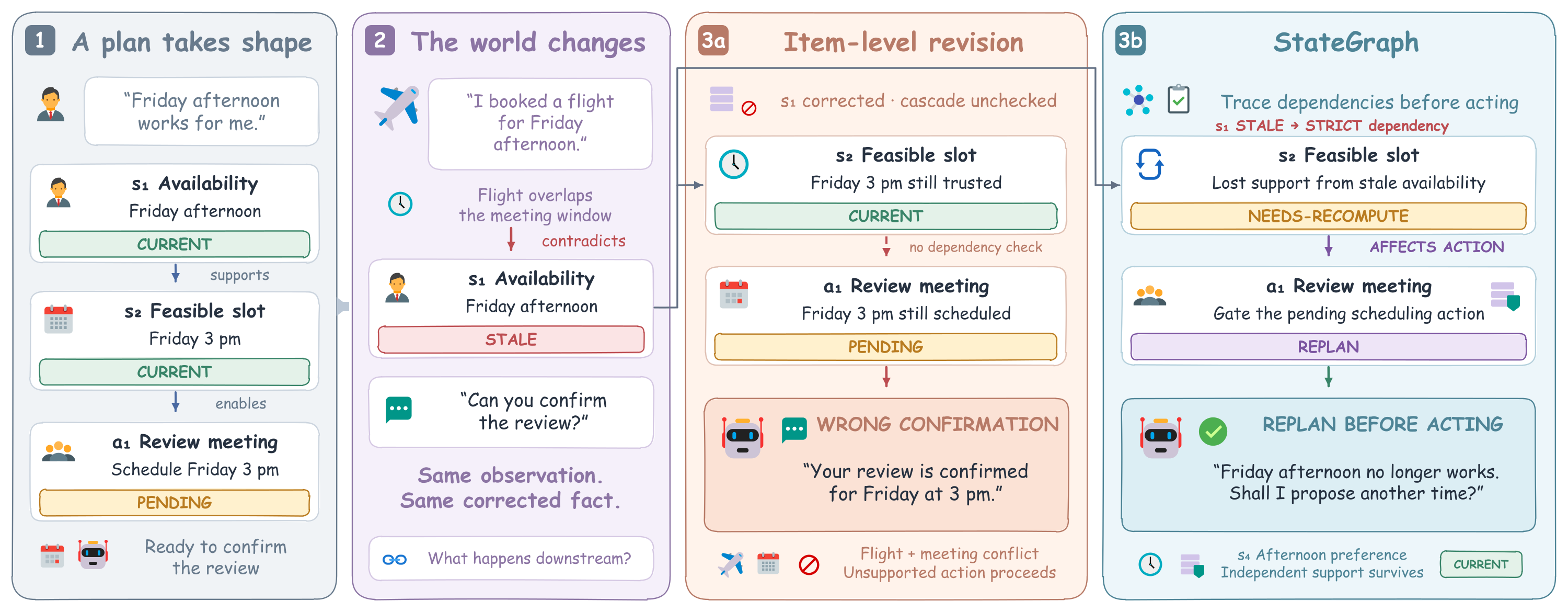}
  \caption{A state change is local, but its consequences are not. The new
  observation contradicts only $s_1$. \textbf{Left:} a system that revises
  only the contradicted memory marks $s_1$ stale but retains $s_2$, permitting the
  pending action $a_1$ to confirm an infeasible meeting.
  \textbf{Right:} StateGraph propagates along verified dependency edges and
  returns the pending action for replanning.}
  \label{fig:teaser}
\end{figure*}

Long-lived agents must update persistent state as availability, tasks,
permissions and bookings change. Existing systems consolidate facts
\citep{chhikara2025mem0}, link notes \citep{xu2025amem}, page context
\citep{packer2023memgpt}, track temporal validity \citep{rasmussen2025zep}
or adjudicate state conflicts with propagation-aware search
\citep{chao2026stale}. These systems address
consolidation, temporal validity and state conflicts, but do not necessarily
identify which downstream conclusions lose support after an upstream revision.

Consider states $s_1$: the user is available Friday afternoon, and $s_2$: a
3pm review is feasible, with pending action $a_1$: schedule the review.
The user then mentions a Friday-afternoon flight, which invalidates $s_1$
without directly contradicting $s_2$, leaving $s_2$ unsupported but still
accepted, so item-level revision can let $a_1$ proceed (Figure~\ref{fig:teaser}).
\textbf{A state change is local, but its consequences need not be.}
Knowledge-editing benchmarks show a similar gap between recalling a changed
fact and reasoning about its implications
\citep{zhong-etal-2023-mquake,cohen-etal-2024-evaluating}.

We therefore formulate \textbf{dependency-aware state revision}: identify
changed states and the downstream beliefs, plans and actions requiring
revision. Three challenges arise. \emph{Discovery} must recover dependencies
that are rarely stated explicitly; \emph{verification} must distinguish support
dependence from semantic similarity; and \emph{propagation} must revise
consequences without invalidating weakly dependent states, which should be
flagged instead, or independently supported ones.

StateGraph addresses these challenges in turn. Bounded candidate discovery
proposes dependency pairs from structural, provenance and linguistic cues
rather than comparing all state pairs. Its core design is counterfactual
verification: each candidate is tested by asking whether the dependent state
would survive the loss of its premise, which grades the edge as \STRICT{},
\WEAK{} or \NODEP{} instead of accepting topical association as dependence.
Propagation then follows verified edges and invalidates a dependent state
only when a \STRICT{} dependency fails and no independent support path
survives, flagging weakly dependent states as \UNCERTAIN{}.

Existing benchmarks evaluate recall, temporal reasoning and
obsolete-information avoidance
\citep{wu2025longmemeval,uddin2026memora,hu2025memoryagentbench}, but lack
joint labels for dependencies, downstream targets, unaffected states and
action consequences. To test the design, we introduce StateChangeBench, a
controlled benchmark of 50 author-constructed cases that provides these
labels. On it, StateGraph's Impact-F1 margin over our CUPMem
reimplementation is 6.9 percentage points at depth~1 and 26.0 at depth~3+,
and on public memory benchmarks its point estimates are comparable or higher
on most datasets. These results provide initial evidence that explicit
dependency tracking can improve selective memory revision when upstream
facts change; establishing this beyond a small controlled benchmark requires
larger independent evaluations.

We make four contributions.
\begin{enumerate}[noitemsep,topsep=3pt,leftmargin=*]
  \item We identify how local state changes invalidate downstream beliefs, plans and actions through latent dependencies without directly contradicting them.
  \item We introduce StateGraph, whose main methodological contribution is recovering support dependencies from interaction histories and verifying them counterfactually; its propagation stage applies established dependency-directed revision with explicit support and scope checks rather than invalidating every reachable state.
  \item We introduce \textbf{StateChangeBench}, a controlled 50-case benchmark annotating dependency structure, cascade depth, hard negatives and action consequences, with metrics for dependency recovery, multi-hop impact and over-/under-invalidation.
  \item We evaluate StateGraph against eight memory baselines and two ported-mechanism variants, with ablations isolating verification, alternative support, conservative revision and propagation (\S\ref{sec:exp}).
\end{enumerate}

\section{Related Work}
\label{sec:related}

\paragraph{Vector and agentic memory.}
RAG couples generation with retrieved non-parametric evidence
\citep{lewis2020rag}, and search-augmented reasoning can learn how much
retrieved evidence to acquire and expose \citep{xiong2026adaptive}. Agent memory adds reflection \citep{park2023generative},
recall-based decay \citep{zhong2024memorybank}, consolidation
\citep{chhikara2025mem0}, linked notes \citep{xu2025amem} and context paging
\citep{packer2023memgpt}. MemoryOS organizes hierarchical storage and updating
\citep{kang2025memoryos}; LightMem combines input compression with offline
consolidation \citep{fang2026lightmem}. DIVE accumulates persistent
natural-language skills and revises them from verifier feedback
\citep{xiong2026dive}; such revision targets reusable procedures rather than
factual states and their consequences. StateGraph focuses on whether
new evidence removes the warrant for downstream states.

\paragraph{Graph-based memory.}
HippoRAG and HippoRAG 2 combine knowledge graphs with Personalized PageRank
for relational retrieval \citep{gutierrez2024hipporag,gutierrez2025hipporag2}.
StructMem preserves event-level bindings and consolidates cross-event
connections \citep{xu2026structmem}. Temporal graphs encode semantic
relations and validity intervals, excluding superseded facts at retrieval
time \citep{rasmussen2025zep}. SWAP represents reasoning steps as an
entailment graph whose intermediate steps can be verified
\citep{xiong2025deliberate}, but the graph is built within a single reasoning
episode rather than maintained across a changing memory. Retrieving related
states and determining which states lose support after a change are distinct
tasks.

\paragraph{State-aware memory.}
STALE \citep{chao2026stale} evaluates implicit conflict: later observations
invalidate earlier memories without explicit negation. CUPMem combines
state consolidation and propagation-aware search. MCB evaluates
whether an agent reusing information should persist it, use it only in
context, re-verify it or ask for clarification \citep{li2026remember}; we
instead ask which downstream states require revision once an upstream state
changes. Our direct revision
(\S\ref{sec:method:direct}) follows write-time adjudication, while
dependency verification determines downstream support.

\paragraph{Knowledge editing and ripple effects.}
Multi-hop editing benchmarks distinguish recalling edits from propagating
their consequences \citep{zhong-etal-2023-mquake,cohen-etal-2024-evaluating}.
These benchmarks evaluate the propagation of triple edits over knowledge
graphs; we recover
latent dependencies from unstructured observations. Parametric editing methods
\citep{meng2022rome,meng2023memit} target model parameters rather than memory states.

\paragraph{Questionable assumptions in QA.}
QA benchmarks study questions with questionable premises
\citep{kim-etal-2023-qa,yu-etal-2023-crepe}. Stale premises are a temporal
instance: a premise was valid earlier but has since changed. Memory records
when it ceased to hold, enabling evidence-backed correction.

\paragraph{Truth maintenance and belief revision.}
Truth maintenance retracts unsupported beliefs \citep{doyle1979tms,dekleer1986atms};
belief revision formalizes change \citep{agm1985}; incremental view
maintenance propagates updates \citep{gupta1995views}. These systems receive
explicit dependencies. StateGraph discovers and counterfactually verifies
dependencies latent in interaction
(\S\ref{sec:method:discovery}--\S\ref{sec:method:verify}), then applies
dependency-directed propagation.

\section{Dependency-Aware State Revision}
\label{sec:problem}

\subsection{State Dependency Graph}

We model memory at time $t$ as a directed graph $G_t=(V_t,E_t)$ whose nodes
partition into evidence $X_t$ (observations, messages, tool outputs), states
$S_t$ and pending actions $A_t$. A state is a tuple
\begin{equation}
s = (e,\; a,\; v,\; \tau,\; \kappa,\; \pi,\; c,\; \sigma),
\end{equation}
where $e$ is an entity, $a$ an attribute, $v$ a value, $\tau$ a temporal scope,
$\kappa$ a condition scope, $\pi \subseteq X_t$ the supporting evidence, $c$ a
confidence, and $\sigma$ a status drawn from
\begin{equation}
\begin{aligned}
\sigma \in \{&\CURRENT,\ \STALE,\ \UNCERTAIN,\\
             &\HISTORICAL,\ \RECOMPUTE\}.
\end{aligned}
\end{equation}
Only \CURRENT{} states may be asserted as facts. \STALE{} marks loss of
current warrant; \HISTORICAL{} retains a previously valid version.
\UNCERTAIN{} marks unknown validity, to be
flagged rather than asserted; \RECOMPUTE{} marks a derived conclusion whose
inputs changed.

Edges are typed. \textsc{supports} links evidence to a state it licenses;
\textsc{updates} and \textsc{invalidates} link a newer state to one it
supersedes or refutes; \textsc{derived-from} links premise states to a conclusion
inferred from them; \textsc{depends-on} links a prerequisite state to a state
whose validity rests on it; \textsc{affects-action} links a state to a pending action for which
it is a precondition. The last three are \emph{dependency edges}, and are the
object of this paper.

\subsection{Two Distinct Tasks}

Let $x_t$ be new evidence. \textbf{Direct revision} identifies the states whose
validity $x_t$ itself settles:
\begin{equation}
D_t = \{\, s \in S_t \;:\; x_t \text{ updates or contradicts } s \,\}.
\end{equation}
This is the task existing systems address. \textbf{Cascade revision} takes the
resulting invalidated set $I_t \subseteq D_t$ and asks which \emph{other}
states or pending actions require revision:
\begin{equation}
\mathrm{Affected}(I_t) \subseteq (S_t\cup A_t) \setminus D_t .
\end{equation}

The central difficulty is that $\mathrm{Affected}(I_t)$ is \emph{not} the set
$R(I_t)$ of states reachable from $I_t$ over dependency edges:
\begin{equation}
\mathrm{Affected}(I_t) \;\subseteq\; R(I_t),
\label{eq:reach}
\end{equation}
with strict inclusion when independent support or scope restrictions leave
reachable nodes CURRENT. Weak dependencies require uncertainty handling,
not wholesale invalidation. Cascade revision must identify the nodes
requiring revision rather than invalidate all of $R(I_t)$.

\subsection{A Consistency Property}
\label{sec:problem:closure}

Gold labels assess revision correctness; conditional invalidation closure
checks consistency given dependency and support judgments. A \STRICT{} edge
states that $s_j$ depended on $s_i$ given the evidence available when the edge
was verified (\S\ref{sec:method:verify}); it does not rule out support that
arises later. A node has \emph{alternative support} when CURRENT,
scope-compatible evidence or states independently license its operative
claim after removal of the failed path. Self-support, failed-path evidence
and circular justification are excluded; reachability alone does not make
otherwise independent support ineligible.

\begin{definition}[Conditional Invalidation Closure]
\label{def:closure}
A memory state satisfies conditional invalidation closure if, for every pair
$s_i \xrightarrow{\textsc{depends-on}} s_j$ that is a \STRICT{} dependency
(\S\ref{sec:method:verify}),
\begin{equation*}
\begin{aligned}
& \sigma(s_i) \neq \CURRENT \\
& \quad \wedge\; \mathrm{AltSupport}(s_j) = \emptyset \\
& \quad \Longrightarrow\; \sigma(s_j) \neq \CURRENT .
\end{aligned}
\end{equation*}
\end{definition}

Weakly dependent states are flagged, while independently supported states
remain valid. \textbf{Closure Violation Rate} (CVR) is the fraction of eligible
pairs violating Definition~\ref{def:closure}, assessed over a common reference
set of dependency pairs;
using only each system's edges would let an empty graph satisfy it vacuously.
CVR measures one-way closure; over-invalidation is measured separately.
Additional non-\CURRENT{} nodes can activate previously inactive closure
requirements. Invalidating all nodes reached in an inferred graph therefore
need not close the common reference graph.

\section{StateGraph}
\label{sec:method}

\begin{figure*}[t]
  \centering
  \includegraphics[width=\textwidth]{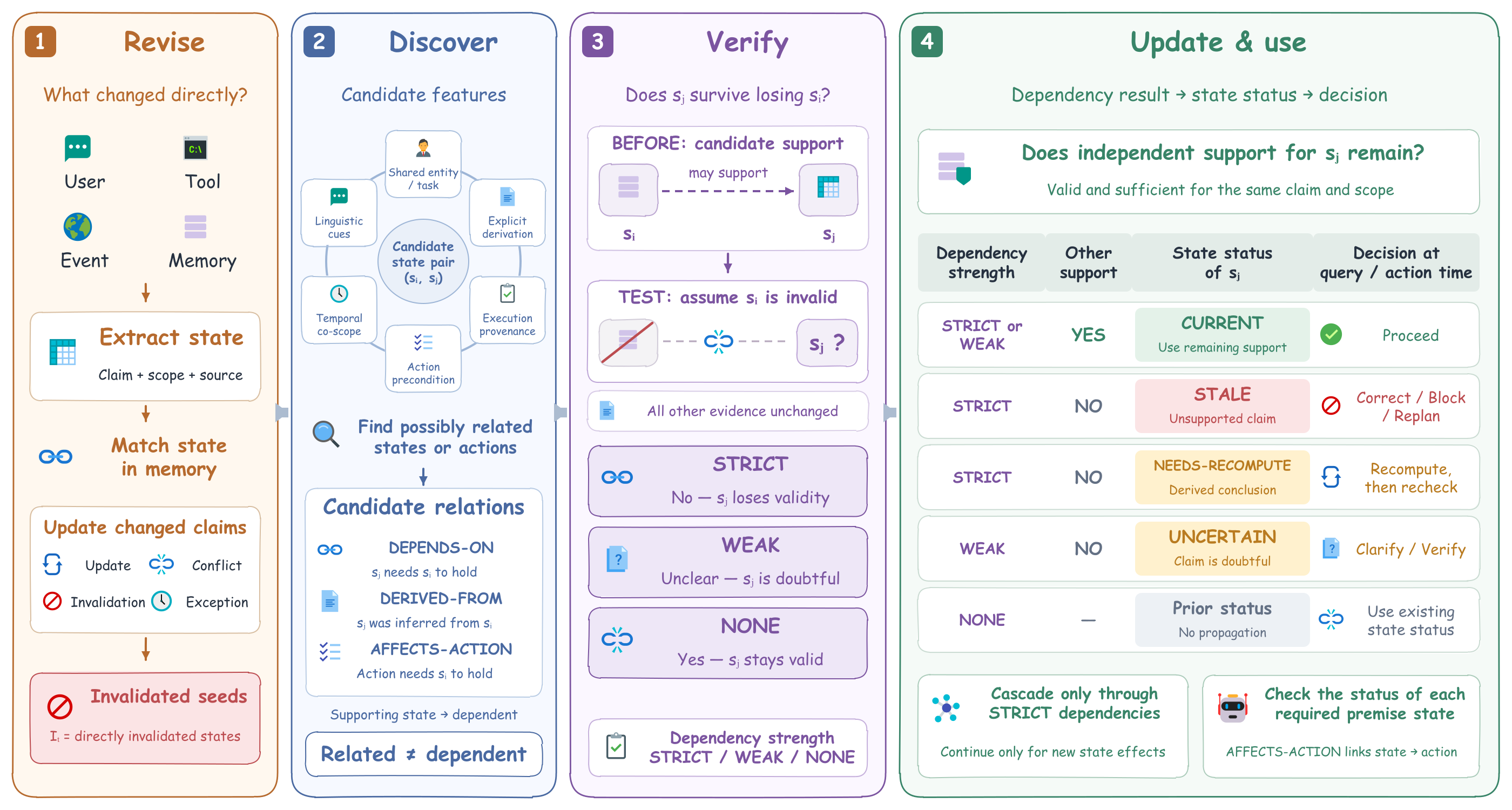}
  \caption{The StateGraph pipeline. \textbf{(A)} Direct revision settles what
  the observation itself contradicts, yielding the seed set $I_t$.
  \textbf{(B)} Dependency reasoning---this paper's contribution---proposes
  candidates from six bounded sources, then verifies each counterfactually, so
  the verified graph is a subgraph of the candidate graph.
  \textbf{(C)} Conservative revision maps each verified edge to a status
  instead of invalidating everything reachable. \textbf{(D)} Premise and action
  gating turns status into behavior. Panels correspond to
  \S\ref{sec:method:direct}--\S\ref{sec:method:premise}.}
  \label{fig:pipeline}
\end{figure*}

StateGraph's main methodological contribution lies in recovering and
verifying support dependencies from interaction histories. Its propagation
stage applies established dependency-directed revision principles
\citep{doyle1979tms} with explicit support and scope checks.
StateGraph incrementally performs direct revision (\S\ref{sec:method:direct}),
candidate discovery (\S\ref{sec:method:discovery}), counterfactual verification
(\S\ref{sec:method:verify}) and conservative propagation
(\S\ref{sec:method:propagate}). Premise-aware use
(\S\ref{sec:method:premise}) gates queries and actions (Figure~\ref{fig:pipeline}).

\subsection{State Representation and Direct Revision}
\label{sec:method:direct}

Observation $x_t$ yields states with type, provenance and temporal/conditional
scope (\S\ref{sec:problem}). Canonical slot identifiers track logical
states; separate identifiers distinguish assertion versions. Matching uses
canonical revision keys and functional identity, meaning that the assertions
describe the same logical state. Each matched pair is classified as
\emph{consistent}, \emph{duplicate}, \emph{update},
\emph{explicit conflict}, \emph{implicit invalidation},
\emph{temporary exception} or \emph{uncertain} (Figure~\ref{fig:pipeline}A).
Updates and conflicts change status while preserving older assertions and
provenance. Scoped overrides distinguish temporary exceptions from general
changes. We adopt the write-time adjudication principle of CUPMem
\citep{chao2026stale} within StateGraph's state schema, canonical matching
and scoped status transitions. This produces the seed set $I_t$;
\S\ref{sec:exp:rq1} compares the implemented systems' direct-revision outcomes.

\subsection{Dependency Candidate Discovery}
\label{sec:method:discovery}

Given $I_t$ and new states, discovery retrieves a bounded pool of historical
states using six structural, revision and provenance sources
(Figure~\ref{fig:pipeline}B1), avoiding exhaustive pairwise comparison:

\begin{itemize}[noitemsep,topsep=3pt,leftmargin=*]
  \item \textbf{Shared entity or task context} --- states about the same
        entity, event or task.
  \item \textbf{Explicit derivation} --- one state was inferred from another in
        an earlier turn.
  \item \textbf{Execution provenance} --- a tool call consumed one state
        as input and produced another.
  \item \textbf{Action preconditions} --- a plan was committed on the basis of
        a state.
  \item \textbf{Temporal co-scoping} --- two states share a constrained window.
  \item \textbf{Linguistic cues} --- \emph{because}, \emph{since},
        \emph{that way}, \emph{so we can}.
\end{itemize}

\paragraph{Directed Relation Typing.}

Discovery returns unordered pairs. Each is examined in both directions,
using $s_i\rightarrow s_j$ for prerequisite-to-dependent edges. Types are
\textsc{depends-on}, where $s_j$'s validity relies on $s_i$;
\textsc{derived-from}, where $s_j$ is inferred from $s_i$; and
\textsc{affects-action}, where $s_i$ is a precondition for action $a_j\in A_t$.
Directional evidence and
endpoint compatibility determine accepted types; candidates with unresolved
direction or insufficient relational evidence are rejected before verification.

Similarity retrieves candidates but never establishes dependency: states
about one meeting may be independent, while a scheduling commitment may
depend on a travel fact with no shared vocabulary. Verification selects
supported dependencies from this candidate pool (\S\ref{sec:exp:rq2}).
Discovery prioritizes recall; verification filters candidates for precision.

\subsection{Counterfactual Dependency Verification}
\label{sec:method:verify}

For each candidate $s_i\rightarrow s_j$, we probe whether $s_j$ survives
antecedent invalidation:

\begin{quote}
\emph{If $s_i$ were no longer a valid current premise, and all other
evidence in the verification context remained unchanged, would $s_j$ still
hold?}
\end{quote}

and label the edge from the answer:

\begin{itemize}[noitemsep,topsep=3pt,leftmargin=*]
  \item \STRICT{} --- $s_j$ does not survive in the verification context. Its
        support there runs through $s_i$.
  \item \WEAK{} --- $s_i$ contributes to $s_j$'s support, but $s_j$ is not
        defeated without it: either its validity becomes undetermined, or
        independent support already in the verification context still
        establishes it.
  \item \NODEP{} --- $s_j$ is unaffected. The states were related, not
        dependent.
\end{itemize}

The verifier rejects judgments inconsistent with their counterfactual outcome.
Accepted \STRICT{} and \WEAK{} edges retain direction, relation type and
dependency strength; \NODEP{} edges do not propagate. The counterfactual
probe is designed to distinguish support dependence from topical relatedness
by testing whether the dependent claim loses its warrant; it also grades the
dependency as \STRICT{} or \WEAK{} and identifies reachable states that
survive antecedent removal (Equation~\ref{eq:reach}).

\paragraph{Strictness and alternative support.}
Verification and revision answer different questions. Verification runs when
an edge is proposed and judges $s_j$ against the support context available to
the verifier at that time. An edge is \STRICT{} only if $s_j$ fails in that
context once $s_i$ is removed; if independent support for $s_j$ is already
present there, $s_j$ holds and the edge is labeled \WEAK{}
(Appendix~\ref{sec:appendix:verify}). \STRICT{} thus means that $s_j$
depended on $s_i$ given the evidence available at verification, not that no
other support can arise later. Invalidating the dependent state therefore
additionally requires that no independent support path survives at revision
time. When $s_i$ later fails,
alternative-support assessment searches the current graph for current,
independent and sufficient evidence or supporting states outside the failed
path, excluding self-support, failed-path evidence and circular justification.
Because support already in the verification context yields \WEAK{}, such
support for a \STRICT{} edge must have arrived after the edge was verified or
lain outside the verifier's bounded context. Strictly dependent states with surviving support
are re-grounded rather than invalidated, preserving hard negatives
(\S\ref{sec:bench}).

\begin{algorithm}[t]
\caption{Conservative Cascade Revision}
\label{alg:verify}
\small
\begin{algorithmic}[1]
\Require verified graph $G_t$, seeds $I_t$
\Ensure revised graph $G'_t$
\State $G'_t\gets\operatorname{copy}(G_t)$

\State $Q \gets \{(s,s,0,[s])\mid s\in I_t\}$
\State $\mathrm{seen}\gets\emptyset$

\While{$Q\neq\emptyset$}
    \State $(s_i,b,d,p)\gets\operatorname{pop}(Q)$

    \ForAll{$(s_j,r,\ell)\in\operatorname{Out}(s_i)$}
        \If{$s_j\in I_t\lor\neg\operatorname{ScopeOK}(s_i,s_j)$}
            \State \textbf{continue}
        \EndIf

        \State $a\gets\operatorname{HasAlt}(s_j,G'_t,I_t,p)$
        \State $\delta\gets
            \operatorname{UpdateRule}(\sigma(s_i),\ell,a,s_j)$

        \If{$\operatorname{IsNew}(s_j,b,\delta,\mathrm{seen},G'_t)$}
            \State $\operatorname{Apply}(G'_t,s_j,\delta)$
            \State $\mathrm{seen}\gets
                \mathrm{seen}\cup\{(s_j,b,\delta)\}$

            \If{$\operatorname{ShouldProp}(\delta)\land\ell=\STRICT$}
                \State $\operatorname{push}(Q,(s_j,b,d+1,p\mathbin{+}[s_j]))$
            \EndIf
        \EndIf
    \EndFor
\EndWhile

\State \textbf{return} $G'_t$
\end{algorithmic}
\end{algorithm}

\subsection{Conservative Cascade Revision}
\label{sec:method:propagate}

Breadth-first propagation from $I_t$ (Algorithm~\ref{alg:verify}) follows
verified edges and Table~\ref{tab:revision-rule}. Each step checks strength,
independent support, state type and temporal/conditional scope; only new
downstream consequences trigger further traversal. Marking weakly supported
states \UNCERTAIN{} rather than \STALE{} parallels confidence-gated
selective prediction, which abstains on low-confidence outputs instead of
committing to them \citep{zheng2026reliablefinancialnamedentity}. Here $b$ identifies the root and $p$ the failed path. IsNew requires an unseen root/node/effect and a new status or support-provenance effect; weaker effects do not overwrite stronger ones.

\par\medskip\noindent
\begingroup
  \centering\small
  \resizebox{\columnwidth}{!}{
  \begin{tabular}{llc}
    \toprule
    \textbf{Edge} & \textbf{Alt.\ support} & \textbf{Consequence} \\
    \midrule
    \STRICT  & none      & \STALE{} / \RECOMPUTE \\
    \STRICT  & survives  & \CURRENT{} (re-grounded) \\
    \WEAK    & none      & \UNCERTAIN{} / Downweight \\
    \WEAK    & survives  & \CURRENT \\
    \NODEP & not assessed & No propagation \\
    Invalid action premise & none & Block / Replan \\
    \bottomrule
  \end{tabular}
  }
  \par
  \captionof{table}{Conservative revision rule, shown graphically in
  Figure~\ref{fig:pipeline}C. Derived conclusions that lose their supporting
  inputs are no longer \CURRENT{} and require re-derivation before reuse;
  \RECOMPUTE{} explicitly records this obligation, while \STALE{} records
  loss of current warrant. Past-valid states are retained as \HISTORICAL{}.}
  \label{tab:revision-rule}
\endgroup
\par\medskip

\subsection{Premise- and Action-Aware State Use}
\label{sec:method:premise}

At use time, StateGraph classifies requests as current-state queries,
historical queries or actions; retrieves states and dependency context;
extracts presuppositions; and checks them against $G_t$ (Figure~\ref{fig:pipeline}D).
\CURRENT{} premises are accepted. \STALE{} premises, or \HISTORICAL{} ones
requiring current validity, trigger correction with the revised state and
the evidence establishing the change.
\UNCERTAIN{} and \RECOMPUTE{} premises trigger clarification or re-derivation.
Actions follow \textsc{affects-action} edges: stale preconditions return a
plan for revision with its invalidation path (Figure~\ref{fig:teaser}).

\paragraph{Complexity.}

Exhaustive adjudication requires $O(|S_t|)$ model calls per observation and
$O(|S_t|^2)$ for pairwise cascade reasoning. With candidate set $\mathcal{C}_t$,
StateGraph uses $O(|\mathcal{C}_t|)$ relation/verification judgments and
$O(|V_{\mathrm{reach}}|+|E_{\mathrm{reach}}|)$ traversal per root, excluding independent-support
assessments. Per cascade case, it uses 12.5K tokens/92.0s, versus 3.5K/45.0s
for vector RAG and 90.6K/340.6s for CUPMem. Full costs appear in
Appendix~\ref{sec:appendix:results}.

\section{StateChangeBench}
\label{sec:bench}

\subsection{What Existing Benchmarks Do Not Isolate}

Existing benchmarks measure recall, retrieval, temporal reasoning, abstention,
environment experience and obsolete-information avoidance
\citep{maharana-etal-2024-evaluating,wu2025longmemeval,wu2026longmemevalv2,uddin2026memora,hu2025memoryagentbench},
without the dependency and downstream-revision labels needed to assess cascades.
Episode-level protocols for workflow agents score state continuity, evidence
traceability and escalation decisions rather than one-shot outputs
\citep{yao2026workflowawarebenchmarkinghealthcarenlp}, and task success can
diverge from the behavior a system is meant to support, so the two should be
reported separately \citep{yao2026beyond}. StateChangeBench likewise scores
post-revision states and downstream actions separately from answer accuracy.

\subsection{Case Design}

Each case includes history, pre-existing states and dependencies, new evidence,
and a query or pending action. Gold labels specify every state's post-revision
status, invalidation depth/path,
\STRICT{}/\WEAK{} strength, surviving-support conditions and action
(\emph{proceed}, \emph{revise-plan}, \emph{ask-clarification}). Strata cover
depth~0--3+, implicit/weak dependencies, alternative support, temporary
exceptions, historical states and action consequences.

\paragraph{Hard negatives.}
Weak dependencies, surviving support and scope-narrowing exceptions test
selective revision against indiscriminate invalidation. StateChangeBench has
50 cases: 10/15/15/10 at depth~0/1/2/3+, including 17 hard negatives (34\%)
and 36 action-consequence cases (72\%). We report over-invalidation alongside
recall; Appendix~\ref{sec:appendix:bench}, Table~\ref{tab:bench-stats},
provides the specification and distribution.

\subsection{Construction and Quality Control}

Cases are seeded from interaction data \citep{wu2025longmemeval} and organized
under a stratum specification. Human annotators label post-update state status,
cascade depth and dependency type. Agreement on a double-annotated subset,
whose size was not recorded, is reported separately for these dimensions, with Cohen's
$\kappa$ values of 0.7915, 1.0000 and 0.5583, respectively. Dependency paths,
downstream revision targets and preservation conditions are specified in
Appendix~\ref{sec:appendix:bench}.

\section{Experiments}
\label{sec:exp}

We evaluate direct revision (RQ1), dependency recovery (RQ2), cascade revision
(RQ3) on StateChangeBench, and report a supplementary evaluation on public
long-term memory benchmarks (RQ4). Action-decision results are not reported
in this version (see Limitations).

\subsection{Setup}
\label{sec:exp:setup}

\paragraph{Baselines.}
Baselines are vector RAG \citep{lewis2020rag}, time-decay RAG, summary memory, Mem0
\citep{chhikara2025mem0}, Zep/Graphiti \citep{rasmussen2025zep},
A-MEM \citep{xu2025amem}, CUPMem \citep{chao2026stale} and Full Context.
We reimplement CUPMem within the shared evaluation framework. Reported token usage
covers the complete memory-processing and response pipeline: on cascade
revision, CUPMem uses 90.6K tokens per case versus StateGraph's
12.5K (approximately 7.2$\times$), distinct from the matched answer-time
context budget.

\paragraph{Ported-mechanism condition.}
Mem0 and CUPMem augmented with cascade revision isolate the mechanism's
contribution while retaining their extraction and retrieval.

\paragraph{Protocol.}
The reported experiments use DeepSeek-v4-Flash. StateChangeBench results pool
counts across repeated evaluation runs per configuration
(Appendix~\ref{sec:appendix:metrics}); the number of runs per configuration
was not preserved in our retained records. Model-request settings and answer-time evidence
limits are specified in Appendix~\ref{sec:appendix:reproducibility};
reported tokens and latency cover all LLM calls per case. StateChangeBench evaluation covers
all 50 cases and depth strata (0, 1, 2, 3+), including ablations
(\S\ref{sec:exp:ablation}). Thinking is disabled. Configuration-specific differences are described
alongside the corresponding experiments. Implementation settings and answer-context limits are detailed in Appendix~\ref{sec:appendix:reproducibility}. Prompt templates appear in
Appendix~\ref{sec:appendix:prompts}.

\subsection{RQ1: Direct State Revision}
\label{sec:exp:rq1}

StateGraph and CUPMem have comparable direct-revision performance across
StateChangeBench depths: direct update accuracy and stale detection are
84.9\%/86.5\% versus 82.7\%/80.3\%, respectively. STALE State Resolution
is 75.4\% versus 72.1\% \citep{chao2026stale}. The stale-detection gain is
6.2 percentage points. RQ1 establishes each system's direct-revision capability
as context for RQ3, which evaluates downstream cascades. Full comparisons
and costs appear in
Appendix~\ref{sec:appendix:results}, Table~\ref{tab:direct_revision}.

\subsection{RQ2: Dependency Recovery}
\label{sec:exp:rq2}

On all 50 StateChangeBench cases, candidate recall, typing recall, verification
F1 and final edge F1 are 62.5\%, 47.0\%, 55.5\% and 61.7\%, respectively
(Appendix~\ref{sec:appendix:results}, Table~\ref{tab:dependency-recovery}).
These measure discovery, typing and verification separately from cascade revision.
Verification F1 summarizes the verification stage, whereas Edge F1 summarizes
final dependency recovery; the two values are not successive scores on a
single prediction task. Candidate and type recall use gold dependency edges
across all 50 cases as their denominator, and Edge F1 compares the final
verified edge set with them. Impact-F1 (RQ3) instead scores predicted
downstream invalidations against gold propagated-invalidation targets in the
40 depth~1--3+ cases.
The two are linked but not interchangeable: an edge error lowers Impact-F1
only if it changes some state's predicted status, so edges ending in states
that stay \CURRENT{}, such as hard negatives, count toward Edge F1 but not
Impact-F1, and Impact-F1 can therefore exceed Edge F1.

\subsection{RQ3: Cascade Revision}
\label{sec:exp:rq3}

We evaluate all 50 StateChangeBench cases. Cascade Impact-F1 is reported on the 40 depth~1--3+ cases, excluding the 10 depth~0 cases. All F1 in Table~\ref{tab:main} is the case-count-weighted mean $(15F_{D1}+15F_{D2}+10F_{D3+})/40$. We also measure impact-set precision and recall
against gold invalidations, plus over-/under-invalidation and CVR
(\S\ref{sec:problem:closure}), with depth-wise results in Table~\ref{tab:main}.
Impact-F1 counts only gold propagated-invalidation targets, excluding directly
revised roots; uncertainty, downweighting and scoped overrides are assessed
separately, and CVR uses strict \textsc{depends-on} reference pairs.
Impact-set and closure formulas appear in Appendix~\ref{sec:appendix:metrics}.

\begin{figure}[t]
    \centering
    \includegraphics[width=0.80\linewidth]{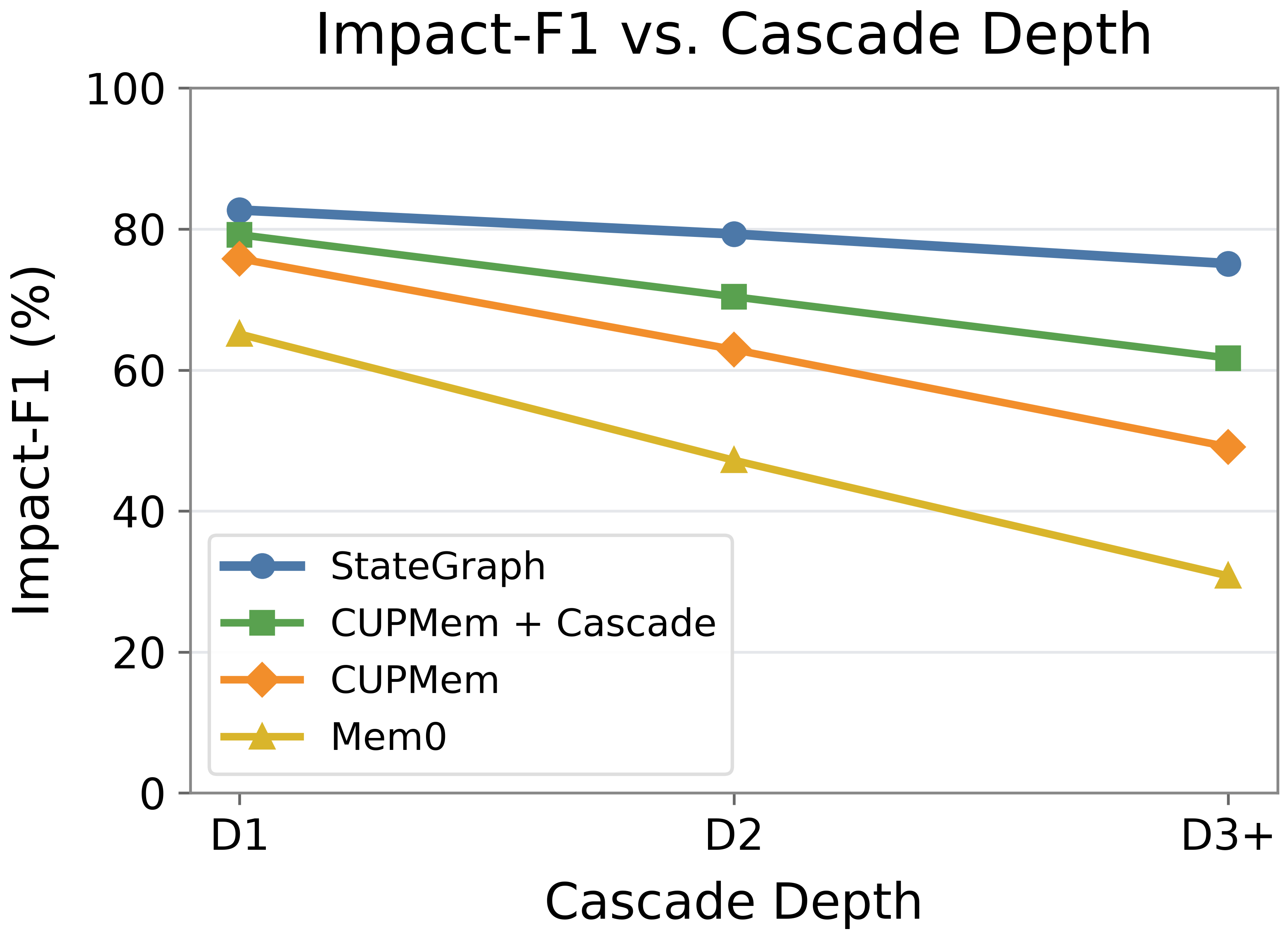}
    \caption{Impact-F1 versus cascade depth on StateChangeBench (15/15/10 cases at depth 1/2/3+). All methods degrade on deeper chains; in these strata, StateGraph degrades least and its margin over the baselines is larger at greater depth.}
    \label{fig:depth}
\end{figure}
\par\medskip\noindent
\begin{minipage}{\columnwidth}
\centering
\footnotesize
\setlength{\tabcolsep}{2.5pt}
\resizebox{\linewidth}{!}{%
\begin{tabular}{lcccc}
\multicolumn{5}{c}{\textbf{Cascade-depth performance}} \\
\toprule
Method & D1 F1 $\uparrow$ & D2 F1 $\uparrow$ & D3+ F1 $\uparrow$ & All F1 $\uparrow$ \\
\midrule
Vector RAG & 20.8 & 10.6 & 5.4 & 13.1 \\
Time-decay RAG & 24.6 & 13.4 & 7.2 & 16.1 \\
Summary Memory & 61.8 & 43.7 & 27.9 & 46.5 \\
Mem0 & 65.1 & 47.2 & 30.8 & 49.8 \\
Zep / Graphiti & 67.4 & 50.3 & 34.6 & 52.8 \\
A-MEM & 68.9 & 51.6 & 35.7 & 54.1 \\
CUPMem & 75.8 & 62.9 & 49.1 & 64.3 \\
Full Context & 72.5 & 57.8 & 42.6 & 59.5 \\
Mem0 + Cascade & 73.4 & 62.1 & 51.0 & 63.6 \\
CUPMem + Cascade & 79.2 & 70.4 & 61.7 & 71.5 \\
\textbf{StateGraph} & \textbf{82.7} & \textbf{79.3} & \textbf{75.1} & \textbf{79.5} \\
\bottomrule
\end{tabular}%
}
\end{minipage}
\par\medskip

Mem0 drops from 65.1\% to 47.2\% F1 at depth~1/2. In the depth strata of
this benchmark, StateGraph's margin over CUPMem is 6.9 percentage points at
depth~1 (82.7\% versus 75.8\%) and 26.0 at depth~3+ (Figure~\ref{fig:depth};
\S\ref{sec:analysis}). With 10--15 cases per stratum, we report this as an
observed pattern on StateChangeBench rather than a general depth trend. Its CVR is 7.6\%, versus CUPMem's 35.1\%;
the other unaugmented baselines range from 38.9\% to 78.2\%. Counterfactual verification and
alternative-support checks limit over-invalidation to 5.2\%.

\par\medskip\noindent
\begin{minipage}{\columnwidth}
\centering
\footnotesize
\setlength{\tabcolsep}{2.5pt}
\resizebox{\linewidth}{!}{%
\begin{tabular}{lccc}
\multicolumn{4}{c}{\textbf{Revision errors}} \\
\toprule
Method & Over-Inv. $\downarrow$ & Under-Inv. $\downarrow$ & CVR $\downarrow$ \\
\midrule
Vector RAG & 31.4 & 73.5 & 78.2 \\
Time-decay RAG & 28.7 & 69.8 & 74.5 \\
Summary Memory & 19.8 & 48.4 & 56.1 \\
Mem0 & 17.2 & 44.1 & 51.8 \\
Zep / Graphiti & 15.6 & 40.8 & 46.3 \\
A-MEM & 14.9 & 39.2 & 44.6 \\
CUPMem & 10.6 & 29.6 & 35.1 \\
Full Context & 12.1 & 33.8 & 38.9 \\
Mem0 + Cascade & 11.2 & 27.0 & 26.5 \\
CUPMem + Cascade & 7.9 & 18.4 & 17.8 \\
\textbf{StateGraph} & \textbf{5.2} & \textbf{10.8} & \textbf{7.6} \\
\bottomrule
\end{tabular}%
}
\captionof{table}{Cascade revision on StateChangeBench. D1--D3+ denote cascade depth; F1 is Impact-F1. All F1 weights D1/D2/D3+ by 15/15/10 cases (40 total), excluding depth~0. Over-/under-invalidation and closure violation rate (CVR) measure revision errors. All values are percentages. Full results, including tokens and latency, appear in Table~\ref{tab:main-full}.}
\label{tab:main}
\end{minipage}
\par\medskip

\subsection{RQ4: Supplementary Evaluation on Public Memory Benchmarks}
\label{sec:exp:rq5}

This supplementary evaluation checks whether adding dependency revision
coincides with lower performance on standard long-term memory tasks. Using
State Answer Accuracy, StateGraph scores 52.0\% on STALE versus CUPMem's
51.7\%, and 53.9\% on LongMemEval \citep{wu2025longmemeval} versus Mem0's
51.3\%. On MemoryAgentBench it scores 36.6\%, below A-MEM's 43.1\% and above
CUPMem's 28.1\%. Its point estimates are comparable to or higher than those
of the evaluated baselines on five of the six benchmarks. We did not run
significance tests, so these results do not establish the absence of
degradation. Appendix~\ref{sec:appendix:results},
Table~\ref{tab:general_long_term_memory}, reports all six benchmarks.

\subsection{Ablations}
\label{sec:exp:ablation}

We remove one mechanism at a time using DeepSeek-v4-Flash on all 50
StateChangeBench cases and depths (Table~\ref{tab:ablation}). Depth-wise results and costs
appear in Appendix~\ref{sec:appendix:results}, Table~\ref{tab:ablation-full}.

\begin{table}[!t]
\centering
\footnotesize
\setlength{\tabcolsep}{2.5pt}
\resizebox{\linewidth}{!}{%
\begin{tabular}{lccc}
\multicolumn{4}{c}{\textbf{Memory revision}} \\
\toprule
Configuration & All F1 $\uparrow$ & Over-Inv. $\downarrow$ & CVR $\downarrow$ \\
\midrule
\textbf{Full StateGraph} & \textbf{79.5} & \textbf{5.2} & \textbf{7.6} \\
\shortstack[l]{w/o Counterfactual\\Verification} & 65.9 & 19.0 & 23.0 \\
\shortstack[l]{w/o Alternative\\Support} & 68.2 & 21.0 & 19.0 \\
\shortstack[l]{w/o Conservative\\Revision} & 62.8 & 25.0 & 28.0 \\
\shortstack[l]{w/o Cascade\\Propagation} & 45.6 & 7.0 & 51.0 \\
LLM-only Revision & 46.7 & 29.0 & 41.0 \\
\bottomrule
\end{tabular}%
}
\caption{Ablations on the full 50-case StateChangeBench using DeepSeek-v4-Flash. All values are percentages. Depth-wise results, under-invalidation and costs appear in Table~\ref{tab:ablation-full}.}
\label{tab:ablation}
\end{table}

\paragraph{Observation 1: verification limits over-invalidation.}
Removing counterfactual verification propagates over all candidate edges:
over-invalidation rises from 5.2\% to 19.0\% and Impact-F1 falls to 65.9
as false edges propagate downstream.

\paragraph{Observation 2: conservative revision improves accuracy.}
Replacing the conservative rule in Table~\ref{tab:revision-rule} with
\STALE{} for every reached dependent node reduces overall Impact-F1 by
16.7 percentage points and raises over-invalidation from 5.2\% to 25.0\%.
The conservative rule preserves independently supported states while
revising unsupported dependents. Removing only alternative-support checks
reduces Impact-F1 by 11.3 percentage points.

\paragraph{Observation 3: the graph supports multi-hop revision.}
Removing cascade propagation reduces D3+ Impact-F1 from 75.1 to 17.9
(Table~\ref{tab:ablation-full}), showing its contribution within StateGraph.
LLM-only revision uses full history without a persistent graph and scores
64.2/25.7 F1 at depth~1/3+. StateGraph achieves 82.7/75.1, gains of
18.5/49.4 points, using 1.67$\times$ the tokens (12.5K versus 7.5K).

\paragraph{Observation 4: cascade revision also helps other memory systems.}
On StateChangeBench, adding cascade revision improves Mem0 and CUPMem
Impact-F1 by 13.8 and 7.2 points, respectively, without changing extraction
or retrieval.

\section{Analysis}
\label{sec:analysis}

\paragraph{Why the margin is larger at greater depth.}
The larger margin at depth~2--3+ is observed on StateChangeBench's depth
strata; the following explanation is consistent with the ablations but is not
separately tested. At depth~1, new evidence often lies near affected states in
topic and time, enabling direct adjudication. At depth~2+, new evidence need
not mention downstream states. Unverified propagation compounds false edges
across hops; counterfactual verification selects supported edges, while
conservative revision preserves independently supported states
(Figure~\ref{fig:depth}; Observations~1--3).

\section{Conclusion}
\label{sec:conclusion}

Long-term agents can fail not only by forgetting or by remembering the wrong
thing, but by continuing to trust conclusions whose grounds have disappeared.
We formulated dependency-aware state revision and introduced StateGraph, which
recovers dependencies, verifies them counterfactually and propagates revisions
selectively, together with StateChangeBench for measuring this capability.
StateGraph addresses downstream loss of support that may remain unresolved
after direct memory revision. Experiments on a small controlled cascade
benchmark suggest benefits from explicit dependency verification and selective
propagation. Larger independent evaluations are needed to establish robustness
across domains, backbones and extended interaction histories. An open problem
is how to verify dependencies without relying on the same model that proposes
them: the current pipeline provides no independent guarantee of detecting
correlated errors across proposal, verification and support assessment.

\section*{Limitations}

The evidence supports dependency-aware revision under the evaluated
conditions: full-benchmark experiments using DeepSeek-v4-Flash, including
all 50 StateChangeBench cases and depth strata. StateChangeBench isolates
cascade reasoning through author-constructed English cases, a written
stratum specification and double annotation of a subset
(Appendix~\ref{sec:appendix:bench}); the broader memory benchmarks provide
complementary coverage. With 50 cases and 10--15 cases per depth stratum,
and without significance tests, the depth-wise differences should be read as
initial evidence.

Our retained records are incomplete. They do not preserve the number of
evaluation runs per configuration or the size of the double-annotated subset,
so these quantities are not reported. For the same reason, this version omits
the action-decision evaluation and the premise-gate ablation: their
aggregation could not be reconstructed from the remaining records. We plan to
rerun these evaluations with complete logging in a future version. These results do not establish performance
across other backbones, languages, multimodal inputs or multi-agent deployments.

Relation typing, counterfactual verification and alternative-support assessment
use the same LLM, so their errors can be correlated. CVR checks closure using
the system's support judgments and is not an independent validation of
those judgments; gold-labeled impact and invalidation metrics provide
complementary outcome evaluation. WEAK edges trigger reassessment without
recursive propagation, restricting cascades to verified STRICT dependencies.
Propagation uses lineage-aware cycle handling; mutually supporting cycles
require separate validation. Verification adds write-time computation; the tables report total per-case
LLM tokens and latency, with costs depending on workload and serving conditions.

\section*{Ethics Statement}

StateGraph addresses outdated premises and unsupported downstream actions
through counterfactual verification, alternative-support checks and
premise-aware gating. Incorrect dependency judgments can nevertheless cause
missed or unnecessary revisions. This version does not evaluate action
decisions, and benchmark results do not certify deployment safety or replace
human authorization for consequential actions.

Evaluation uses existing memory benchmarks and StateChangeBench to assess
memory revision and downstream action behavior. For deployments that store
personal states, memory management should also incorporate data minimization,
access controls, appropriate retention periods, and user correction and deletion
mechanisms for both memories and dependency links.

\bibliography{references}

\clearpage
\appendix
\onecolumn
\addtocontents{toc}{\protect\setcounter{tocdepth}{2}}
\begingroup
\renewcommand{\contentsname}{Appendix Contents}
\tableofcontents
\endgroup
\medskip

\section{Additional Experimental Results}
\label{sec:appendix:results}

This section retains the complete baseline comparisons and diagnostic
breakdowns. All experiments use DeepSeek-v4-Flash and evaluate full benchmarks;
the StateChangeBench comparisons and ablations cover all 50 cases. Evaluation coverage and metric scope are distinct: main-table All F1 aggregates only the 40 depth~1--3+ cases.

\subsection{Direct State Revision}
\par\medskip\noindent
\begin{minipage}{\textwidth}
\centering
\resizebox{\textwidth}{!}{
\begin{tabular}{lccccc}
\toprule
\textbf{Method}
& \textbf{SCB Update Acc. $\uparrow$}
& \textbf{SCB Stale Detection $\uparrow$}
& \textbf{STALE State Resolution $\uparrow$}
& \textbf{Ref. Tokens / Case $\downarrow$}
& \textbf{Ref. Latency / Case $\downarrow$} \\
\midrule
Vector RAG       & 22.4 & 38.1 & 15.0 & 3.5K & 45.0 \\
Time-decay RAG   & 26.5 & 36.8 & 20.6 & 3.6K & 46.0 \\
Summary Memory   & 70.0 & 68.4 & 20.9 & 8.8K & 100.0 \\
Mem0             & 74.3 & 72.7 & 10.3 & 34.4K & 166.1 \\
Zep / Graphiti   & 76.8 & 74.0 & 6.5 & 40.6K & 220.0 \\
A-MEM            & 78.1 & 76.6 & 10.8 & \textbf{2.6K} & \textbf{38.0} \\
CUPMem           & 82.7 & 80.3 & 72.1 & 90.6K & 340.6 \\
Full Context     & 80.2 & 78.5 & 55.7 & 6.0K & 59.0 \\
\midrule
\textbf{StateGraph}
& \textbf{84.9}
& \textbf{86.5}
& \textbf{75.4}
& 12.5K & 92.0 \\
\bottomrule
\end{tabular}
}
\captionof{table}{Direct state revision performance on StateChangeBench and STALE.
Higher is better for accuracy metrics; lower is better for reference costs. Costs are reproduced from Table~\ref{tab:main-full} as a common StateChangeBench reference, not separate measurements for each task-specific metric.}
\label{tab:direct_revision}
\end{minipage}
\par\medskip

\subsection{Dependency Recovery}
Table~\ref{tab:dependency-recovery} evaluates dependency edges, whereas
Impact-F1 (Table~\ref{tab:main-full}) evaluates downstream invalidations.
Candidate recall, type recall and Edge F1 are computed against the gold
dependency edges of all 50 cases; Impact-F1 is computed against the gold
propagated-invalidation targets of the 40 depth~1--3+ cases. An edge error changes Impact-F1 only when it
changes a state's predicted status, so the two need not move together.
\par\medskip\noindent
\begin{minipage}{\columnwidth}
\centering
\small
\begin{tabular}{lcccc}
\toprule
\textbf{Method}
& \textbf{Cand. Rec. $\uparrow$}
& \textbf{Type Rec. $\uparrow$}
& \textbf{Verif. F1 $\uparrow$}
& \textbf{Edge F1 $\uparrow$} \\
\midrule
\textbf{StateGraph}
& 62.5
& 47.0
& 55.5
& 61.7 \\
\bottomrule
\end{tabular}
\captionof{table}{Dependency recovery on the full 50-case StateChangeBench.
Higher is better for all metrics.}
\label{tab:dependency-recovery}
\end{minipage}
\par\medskip

\subsection{Cascade Revision and Computational Cost}
\par\medskip\noindent
\begin{minipage}{\textwidth}
\centering
\resizebox{\textwidth}{!}{
\begin{tabular}{lcccccccc}
\toprule
\textbf{Method}
& \textbf{D1 F1 $\uparrow$}
& \textbf{D2 F1 $\uparrow$}
& \textbf{D3+ F1 $\uparrow$}
& \textbf{All F1 $\uparrow$}
& \textbf{Over-Inv. $\downarrow$}
& \textbf{Under-Inv. $\downarrow$}
& \textbf{CVR $\downarrow$}
& \textbf{Tokens / Latency $\downarrow$} \\
\midrule

Vector RAG
& 20.8 & 10.6 & 5.4 & 13.1
& 31.4 & 73.5 & 78.2
& 3.5K / 45.0s \\

Time-decay RAG
& 24.6 & 13.4 & 7.2 & 16.1
& 28.7 & 69.8 & 74.5
& 3.6K / 46.0s \\

Summary Memory
& 61.8 & 43.7 & 27.9 & 46.5
& 19.8 & 48.4 & 56.1
& 8.8K / 100.0s \\

Mem0
& 65.1 & 47.2 & 30.8 & 49.8
& 17.2 & 44.1 & 51.8
& 34.4K / 166.1s \\

Zep / Graphiti
& 67.4 & 50.3 & 34.6 & 52.8
& 15.6 & 40.8 & 46.3
& 40.6K / 220.0s \\

A-MEM
& 68.9 & 51.6 & 35.7 & 54.1
& 14.9 & 39.2 & 44.6
& \textbf{2.6K / 38.0s} \\

CUPMem
& 75.8 & 62.9 & 49.1 & 64.3
& 10.6 & 29.6 & 35.1
& 90.6K / 340.6s \\

Full Context
& 72.5 & 57.8 & 42.6 & 59.5
& 12.1 & 33.8 & 38.9
& 6.0K / 59.0s \\

\midrule

Mem0 + Cascade
& 73.4 & 62.1 & 51.0 & 63.6
& 11.2 & 27.0 & 26.5
& 41.0K / 190.0s \\

CUPMem + Cascade
& 79.2 & 70.4 & 61.7 & 71.5
& 7.9 & 18.4 & 17.8
& 104.0K / 385.0s \\

\textbf{StateGraph}
& \textbf{82.7}
& \textbf{79.3}
& \textbf{75.1}
& \textbf{79.5}
& \textbf{5.2}
& \textbf{10.8}
& \textbf{7.6}
& 12.5K / 92.0s \\

\bottomrule
\end{tabular}
}
\captionof{table}{Main results on StateChangeBench. All F1 is the 15/15/10 case-count-weighted mean over D1/D2/D3+ (40 cases), excluding depth~0. Higher is better for Impact-F1;
lower is better for over-/under-invalidation, CVR, and computational cost.}
\label{tab:main-full}
\end{minipage}
\par\medskip

\subsection{Generic Long-Term Memory}
\par\medskip\noindent
\begin{minipage}{\textwidth}
\centering
\resizebox{\textwidth}{!}{
\begin{tabular}{lcccccc}
\toprule
\textbf{Method}
& \textbf{LongMemEval $\uparrow$}
& \textbf{LME-V2 $\uparrow$}
& \textbf{MemoryAgentBench $\uparrow$}
& \textbf{Memora $\uparrow$}
& \textbf{STALE $\uparrow$}
& \textbf{LoCoMo $\uparrow$} \\
\midrule
Mem0 & 51.3 & 14.1 & 19.0 & 23.9 & 6.2 & 56.7 \\
Zep / Graphiti & 50.8 & 19.5 & 22.9 & 24.1 & 4.0 & 55.3 \\
A-MEM & 39.5 & 11.7 & \textbf{43.1} & 28.6 & 3.1 & 38.4 \\
CUPMem & 47.3 & 15.2 & 28.1 & 25.5 & 51.7 & 50.8 \\
\textbf{StateGraph} & \textbf{53.9} & \textbf{22.7} & 36.6 & \textbf{31.8} & \textbf{52.0} & \textbf{58.0} \\
\bottomrule
\end{tabular}
}
\captionof{table}{State Answer Accuracy (\%) across six public long-term memory benchmarks. Higher is better.}
\label{tab:general_long_term_memory}
\end{minipage}
\par\medskip

\subsection{Full Diagnostic Ablations}
\par\medskip\noindent
\begin{minipage}{\textwidth}
\centering
\small
\setlength{\tabcolsep}{4pt}
\textbf{Memory revision}
\par\smallskip
\begin{tabular}{lccccccc}
\toprule
Configuration & D1 F1 $\uparrow$ & D2 F1 $\uparrow$ & D3+ F1 $\uparrow$ & All F1 $\uparrow$ & \shortstack{Over-Inv.\\$\downarrow$} & \shortstack{Under-Inv.\\$\downarrow$} & CVR $\downarrow$ \\
\midrule
\textbf{Full StateGraph} & \textbf{82.7} & \textbf{79.3} & \textbf{75.1} & \textbf{79.5} & \textbf{5.2} & 10.8 & \textbf{7.6} \\
\shortstack[l]{w/o Counterfactual\\Verification} & 75.6 & 64.3 & 54.0 & 65.9 & 19.0 & 13.0 & 23.0 \\
\shortstack[l]{w/o Alternative\\Support} & 77.1 & 66.5 & 57.2 & 68.2 & 21.0 & 10.0 & 19.0 \\
\shortstack[l]{w/o Conservative\\Revision} & 74.3 & 60.5 & 49.0 & 62.8 & 25.0 & \textbf{9.5} & 28.0 \\
\shortstack[l]{w/o Cascade\\Propagation} & 71.2 & 38.4 & 17.9 & 45.6 & 7.0 & 58.0 & 51.0 \\
LLM-only Revision & 64.2 & 43.1 & 25.7 & 46.7 & 29.0 & 42.0 & 41.0 \\
\bottomrule
\end{tabular}
\par\medskip
\textbf{Computational cost}
\par\smallskip
\begin{tabular}{lcc}
\toprule
Configuration & Tokens $\downarrow$ & Latency $\downarrow$ \\
\midrule
\textbf{Full StateGraph} & 12.5K & 92.0s \\
\shortstack[l]{w/o Counterfactual\\Verification} & 10.0K & 76.0s \\
\shortstack[l]{w/o Alternative\\Support} & 11.7K & 87.0s \\
\shortstack[l]{w/o Conservative\\Revision} & 12.4K & 91.0s \\
\shortstack[l]{w/o Cascade\\Propagation} & 8.6K & 64.0s \\
LLM-only Revision & \textbf{7.5K} & \textbf{60.0s} \\
\bottomrule
\end{tabular}
\captionof{table}{Ablation results for StateGraph on the
full 50-case StateChangeBench using DeepSeek-v4-Flash.
Higher is better for F1;
lower is better for error rates and computational cost.}
\label{tab:ablation-full}
\end{minipage}
\par\medskip

\subsection{Implementation and Reproducibility}
\label{sec:appendix:reproducibility}

\paragraph{Propagation.}
Cascade traversal has no fixed numerical depth bound. It terminates when
its queue is exhausted, deduplicating source/root/effect lineages and
skipping cycles back to directly invalidated seeds. Only new status effects
trigger further traversal; weak dependencies prompt revalidation without
extending the cascade. Depth in Algorithm~\ref{alg:verify} records traversal
distance rather than imposing a cutoff. Retrieval graph depth 2, beam width
4 and relational maximum hops 3 are separate retrieval settings.

\paragraph{Candidate limits.}
The modular discovery request permits at most 64 unordered endpoint pairs.
Structural candidates are then added and unordered pairs deduplicated before
separate direction/type classification. The 64-pair limit applies to each
model response, not to the combined pool after structural supplementation.

\paragraph{Generation.}
All experiments use DeepSeek-v4-Flash with thinking disabled. Maximum output tokens are 8,192 for fact
proposal; 2,048 each for semantic validation, direct revision and
cross-observation dependency proposal; 4,096 for dependency verification;
and 512 for shared answer generation. These are request ceilings, not
observed completion lengths. Each logical request permits up to three
transport attempts, with SDK retries and structured-output repair attempts
set to zero.

\paragraph{Answer-time evidence.}
Retrieval selects up to ten states, prioritizing query-field coverage and
provenance anchors, then expanding dependency neighbors breadth-first.
Serialized evidence spans are limited to 480 characters each. The final
answer-context helper preserves input order, admits at most ten items and
64,000 characters, truncates the first over-budget item and stops. The
configuration also declares 16,000 tokens; the helper enforces item and
character limits rather than a tokenizer-based hard token cap. These limits
apply only to answer-time context. Tokens / Case sums recorded usage across
all case-level LLM calls, including proposal, validation, revision,
dependency reasoning and answer generation.

\subsection{Impact-Set and Closure Metrics}
\label{sec:appendix:metrics}

Impact-F1 evaluates downstream state invalidation against explicit gold
propagated-invalidation targets, excluding directly revised roots. Uncertainty,
downweighting, scoped overrides and action decisions are assessed separately.
For case $c$ in run $r$, let $\widehat{Y}_{cr}$ and $Y_c$ denote the predicted
and gold downstream invalidation sets. Within each depth stratum $d$, counts
are pooled across cases and runs:
\begin{equation}
\begin{aligned}
\mathrm{TP}_d&=\sum_{r,c\in d}|\widehat{Y}_{cr}\cap Y_c|,\\
\mathrm{FP}_d&=\sum_{r,c\in d}|\widehat{Y}_{cr}\setminus Y_c|,\\
\mathrm{FN}_d&=\sum_{r,c\in d}|Y_c\setminus\widehat{Y}_{cr}|.
\end{aligned}
\end{equation}
Precision and recall are $\mathrm{TP}_d/(\mathrm{TP}_d+\mathrm{FP}_d)$ and
$\mathrm{TP}_d/(\mathrm{TP}_d+\mathrm{FN}_d)$ when their denominators are
positive. Stratum-level Impact-F1 is
\begin{equation}
F_d=\frac{2\mathrm{TP}_d}{2\mathrm{TP}_d+\mathrm{FP}_d+\mathrm{FN}_d}.
\end{equation}
Empty predicted and gold sets contribute zero counts. A stratum with no
predicted or gold targets is reported as not applicable. In both main and
ablation tables, All F1 weights the depth-1, depth-2 and depth-3+ scores by
15, 15 and 10 cases, respectively. Depth-0 cases remain part of the full
benchmark evaluation but are excluded from All F1.

Over-invalidation is the proportion of gold-preserved states incorrectly
invalidated. Under-invalidation is the proportion of gold downstream
invalidation targets that remain uninvalidated. Each rate pools its numerator
and denominator across evaluated cases and runs.

CVR uses a shared reference set $E_{\mathrm{ref},c}$ of strict
\textsc{depends-on} state pairs covered by Definition~\ref{def:closure}.
For each case and run, activated pairs are
\begin{equation}
\begin{aligned}
\Omega_{cr}=\{(i,j)\in E_{\mathrm{ref},c}:{}&
\sigma_{cr}(s_i)\neq\CURRENT,\\
&\mathrm{AltSupport}_{cr}(s_j)=\emptyset\}.
\end{aligned}
\end{equation}
A violation occurs when the dependent remains \CURRENT{}. Aggregate CVR is
\begin{equation}
\mathrm{CVR}=\frac{\sum_{c,r}|\{(i,j)\in\Omega_{cr}:
\sigma_{cr}(s_j)=\CURRENT\}|}{\sum_{c,r}|\Omega_{cr}|}.
\end{equation}
Cases without activated pairs contribute to neither count; CVR is not
applicable when the aggregate denominator is zero. Reference pairs are shared
across systems, while activation depends on each system's resulting state and
support judgments. CVR measures conditional closure; over-invalidation
separately measures unnecessary invalidation.

\section{Prompt Templates}
\label{sec:appendix:prompts}

The templates below specify module instructions, runtime inputs and output contracts.

\subsection{State Extraction}

\begin{prompttemplate}{01 | State Extraction}
SYSTEM
Extract atomic source assertions with exact evidence; no invented facts/actions/revisions. Preserve entity/value, scope, provenance and event roles. Copy explicit derivation/action inputs; separate reported_content/claimant.
CURRENT by default; HISTORICAL/UNCERTAIN require evidence. Negation preserves positive content; absolute dates require a year; speakers require attribution.
INPUT: case_id, observation_id, observation_text; optional speaker_spans, normalization_feedback.
OUTPUT: states[] with entity/attribute/value, scope/event_frame/evidence and:
node_type: FACT|CONDITION|ROLE_ASSERTION|DERIVED_STATE|ACTION|ACTION_PRECONDITION.
claim_mode: DIRECT|REPORTED|ASSUMPTION|INTENT|ADVICE.
polarity: AFFIRMED|NEGATED|UNKNOWN; time: NONE|TEXT|ABSOLUTE.
PROPERTY_IDENTITY/REPORTED_PROPOSITION provide source-grounded projections.
\end{prompttemplate}

\subsection{Direct Revision}

\begin{prompttemplate}{02 | Direct State Revision}
SYSTEM
Match canonical slots/functional keys; revise direct assertions only. Separate scopes coexist. Replacement: UPDATE/MARK_HISTORICAL; same-scope denial/broken guarantee: EXPLICIT_CONFLICT/MARK_STALE. Preserve outside TEMPORARY_EXCEPTION scope.
INPUT: existing_states, incoming_states, new_observation_text.
OUTPUT: decisions[] with incoming_state_id/existing_state_ids, relation/transition, scope_text/evidence_span, reason (<=12 words), confidence; one decision per incoming ID.
relation: NEW_FACT|CONSISTENT|DUPLICATE|UPDATE|EXPLICIT_CONFLICT|IMPLICIT_INVALIDATION|TEMPORARY_EXCEPTION|UNCERTAIN.
transition: KEEP_CURRENT|KEEP_BOTH|MARK_HISTORICAL|MARK_STALE|SCOPED_OVERRIDE|MARK_UNCERTAIN|RECOMPUTE.
No benchmark labels/answers.
\end{prompttemplate}

\subsection{Dependency Candidate Discovery}

\begin{prompttemplate}{03 | Candidate Discovery}
SYSTEM
Propose unordered support/derivation/precondition pairs; no direction/type/strength or topic-only links. Supplied IDs and pair limit only; reason <=8 words; no gold labels.
INPUT: {"states": [...]}
OUTPUT: {"pairs":[{"state_id_a":"<ID>","state_id_b":"<ID>","reason":"<reason>"}]}
\end{prompttemplate}

\subsection{Dependency Direction and Type}

\begin{prompttemplate}{04 | Direction and Relation Type}
SYSTEM
Store prerequisite -> dependent. DEPENDS_ON: support; DERIVED_FROM: named input; AFFECTS_ACTION: pending-action prerequisite. Respect types/claims and participation IDs (COMMITMENT supplies; ASSIGNMENT requires).
Hold other assertions fixed; test both directions. Neither: NONE; both: ambiguous. No topical/transitive links or usefulness-only invalidation.
INPUT: nodes {A,B}, source_semantic_links, candidate_provenance.
OUTPUT: prerequisite/dependent A|B|NONE, relation DEPENDS_ON|DERIVED_FROM|AFFECTS_ACTION|NONE, counterfactual booleans, confidence, <=2 A/B references, reason.
\end{prompttemplate}

\subsection{Counterfactual Dependency Verification}
\label{sec:appendix:verify}

\begin{prompttemplate}{05 | Counterfactual Verification}
SYSTEM
Remove P as CURRENT warrant; hold records/other assertions fixed. Test D validity/eligibility. Independent support only; no circular/downstream warrants or invented requirements.
STRICT_DEPENDENCY: grounded bridge + FAILS; WEAK_DEPENDENCY: contributory bridge + HOLDS/UNDETERMINED; NO_DEPENDENCY: no bridge. FAILS=lost warrant; HOLDS=independent support; UNDETERMINED=reassess.
INPUT: tested_claims, validity_contract, condition_bridge, proposed_relation, verification_epoch, source_evidence, remaining_evidence/support_context, intervention.
OUTPUT: strength, outcome, P/D/O references (positive decisions cite P/D), reason <=120 characters; honor STRICT restrictions.
INFLUENCE: influence_present, component VALIDITY|TIMING|PURPOSE|METHOD|VALUE|NONE, references/reason.
\end{prompttemplate}

\subsection{Alternative Support Detection}

\begin{prompttemplate}{06 | Independent Alternative Support}
SYSTEM
Find CURRENT independent evidence fully supporting target_claim in the same scope after failed-path removal; alternative derivations are allowed.
Exclude self, failed-path/dependent and historical evidence. Repetition/missing edges do not prove independence.
INPUT: target_claim, dependent, failed roots/path, blocked IDs, eligible_sources, excluded_sources.
OUTPUT: support_ids, independence_established, fully_supported, reason <=120 characters. Unsupported: false/[]; no invented evidence.
\end{prompttemplate}

\subsection{Premise Checking and Action Gating}

\begin{prompttemplate}{07 | Premise Checking and Action Gating}
INTENT
From query wording choose CURRENT_QUERY|HISTORICAL_QUERY|ACTION_REQUEST; future use of old plans is not historical.
INPUT: query. OUTPUT: intent/reason.

REFERENCES
Select supplied requested_ids (asked about), premise_ids (presupposed CURRENT), action_ids (requested actions); respect intent without guessing.
INPUT: query, intent, typed/scoped nodes with aliases/evidence.
OUTPUT: requested_ids, premise_ids, action_ids, reason <100 characters.
Code checks dependency/status/scope/action eligibility; no gate prompt.
\end{prompttemplate}

\subsection{Answer and Action Generation}

\begin{prompttemplate}{08 | Premise-Aware Answer and Action Generation}
SYSTEM
Use CURRENT evidence within applicability_scopes; historical evidence only for historical queries. Correct stale premises; obey original_action_execution_allowed; no blocked execution/invented resources.
CURRENT inputs and verified_recomputation_results govern values/units. Sufficient inputs: COMPUTATION; otherwise REASSESSMENT_REQUIRED/null. Reassess downweighted claims; missing confirmation is not unavailability.
INPUT: query, runtime_gate, verified_recomputation_results, evidence_aliases.
OUTPUT: answer, action_decision NO_ACTION|EXECUTE|REVISE_PLAN|BLOCK|CLARIFY, CURRENT used_evidence_refs, <=5 replanned_actions; respect runtime restrictions.
REVISE_PLAN proposes grounded alternatives; BLOCK only blocks; CLARIFY requests inputs. No execution.
Recomputation: COMPUTABLE|INSUFFICIENT_INPUT, expression/input_refs/reason; restricted decimal arithmetic.
\end{prompttemplate}

\paragraph{Shared benchmark answer generation.}
\begin{prompttemplate}{09 | Shared Answer Generation}
SYSTEM
Return the requested fact/value from context only, without explanation; state insufficiency when unsupported.
USER
Context: {retrieved_context}
Question: {question}
\end{prompttemplate}

\paragraph{Implementation sources.}
The modular sources are \texttt{run\_state\_extraction.py},
\texttt{revision\_v3.py}, \texttt{dependency\_candidates\_v3.py},
\texttt{relation\_typing\_v3.py},
\texttt{counterfactual\_verification\_v3.py},
\texttt{alternative\_support\_v3.py}, \texttt{premise\_gate\_v3.py}
and \texttt{recomputation\_v3.py}. The shared answer configuration is
\texttt{shared\_answer\_generation.yaml}.

\section{Benchmark Construction Details}
\label{sec:appendix:bench}

\subsection{Benchmark Statistics}

\par\medskip\noindent
\begin{minipage}{\textwidth}
  \centering
  \small
  \begin{tabular}{lrrr}
    \toprule
    \textbf{Stratum} &
    \textbf{Cases} &
    \textbf{Share} &
    \textbf{Mean cascade size} \\
    \midrule
    Hard negatives        & 17 & 34\% & 1.882 \\
    Depth 0               & 10 & 20\% & 0.000 \\
    Depth 1               & 15 & 30\% & 1.000 \\
    Depth 2               & 15 & 30\% & 2.000 \\
    Depth 3+              & 10 & 20\% & 3.000 \\
    Implicit dependency   & 7  & 14\% & 1.857 \\
    Weak dependency       & 10 & 20\% & 1.900 \\
    Alternative support   & 4  & 8\%  & 1.750 \\
    Temporary exception   & 6  & 12\% & 1.500 \\
    Historical state      & 2  & 4\%  & 2.500 \\
    Action consequence    & 36 & 72\% & 1.694 \\
    \midrule
    Status $\kappa$          & \multicolumn{3}{c}{0.7915} \\
    Depth $\kappa$           & \multicolumn{3}{c}{1.0000} \\
    Dependency-type $\kappa$ & \multicolumn{3}{c}{0.5583} \\
    \bottomrule
  \end{tabular}
  \captionof{table}{Statistics of the StateChangeBench benchmark.}
  \label{tab:bench-stats}
\end{minipage}
\par\medskip

\subsection{Stratum Specification}

Case authors work from the following specification and are not shown the
method. Each case must be solvable from the history alone.

\begin{itemize}[noitemsep,topsep=3pt,leftmargin=*]
  \item \textbf{Depth 0} --- the new observation directly contradicts exactly
        one state; nothing downstream depends on it.
  \item \textbf{Depth 1} --- exactly one state depends on the contradicted
        state, and nothing in the history contradicts that dependent state.
  \item \textbf{Depth 2, 3+} --- as above, chained. Every intermediate state
        must be individually plausible in isolation.
  \item \textbf{Implicit dependency} --- the dependency is never stated
        linguistically and must be inferred from task structure.
  \item \textbf{Weak dependency} --- the antecedent raised the dependent
        state's plausibility without establishing it; the gold status is
        \UNCERTAIN{}, not \STALE{}.
  \item \textbf{Alternative support} --- a second, independent line of evidence
        licenses the dependent state; gold status is \CURRENT{}.
  \item \textbf{Temporary exception} --- the change narrows a scope rather than
        voiding it; the general state survives, the scoped instance does not.
  \item \textbf{Historical state} --- the superseded state remains correct as
        history and must stay retrievable.
  \item \textbf{Action consequence} --- a pending action depends on the
        affected state; the gold action label is not \emph{proceed}.
\end{itemize}

\subsection{Annotation Protocol}

Every gold-invalidated state carries its depth, the dependency path that
reaches it and a \STRICT{}/\WEAK{} label; surviving reachable states carry
their preservation conditions. Human annotators label state status, cascade
depth and dependency type. Cohen's $\kappa$ is reported separately for these
dimensions on a double-annotated subset; the size of this subset was not
recorded.

\subsection{Hard Negatives}

Hard negatives comprise 17 of 50 cases (34\%). Table~\ref{tab:bench-stats}
reports case counts and shares across the benchmark strata.

\section{Case Studies}
\label{sec:appendix:cases}

\subsection{Study Case}

\noindent\textit{Evaluation setting.} This case study reports a StateGraph run using DeepSeek-v4-Flash on StateChangeBench case \texttt{SCB\_006}, with the results compared against its gold annotations.

Case \texttt{SCB\_006} is a D3+ example with structural depth three. Its history records that order \#4471 was addressed to 12 Elm Street, a courier label was printed from that address, a Thursday estimate was based on the label route, and an arrival email was queued for Lena. A handwritten gift note is recorded separately. The new observation says the parcel should not go to Elm Street and requests 88 Oak Avenue instead. The query asks to tell Lena that the parcel will arrive Thursday at the old address.

StateGraph first aligned the new address assertion with the existing address slot. Given the source's explicit correction, direct revision marked the old address stale while retaining the replacement as current. The gold graph specifies three strict links: address to label (\texttt{DEPENDS\_ON}), label to estimate (\texttt{DERIVED\_FROM}), and estimate to queued email (\texttt{AFFECTS\_ACTION}). Relation typing and counterfactual verification accepted these source-grounded links. Propagation assessed each downstream state for independent current support and marked the label, estimate, and queued email stale at depths one through three, matching the gold targets. The unrelated gift note remained current.

At query time, premise-aware gating rejected the old address as current evidence and blocked the original action referencing the stale chain, requiring replanning. The benchmark identifiers are retained below; the implementation distinguishes state nodes from pending actions when applying \texttt{AFFECTS\_ACTION}.

\par\medskip\noindent
\begin{minipage}{\textwidth}
\centering
\small
\captionof{table}{StateGraph revision results using DeepSeek-v4-Flash on SCB\_006, compared with the gold annotations.}
\label{tab:study-case-scb006}
\begin{tabular}{@{}p{0.16\linewidth}p{0.37\linewidth}p{0.40\linewidth}@{}}
\toprule
\textbf{Item} & \textbf{Before correction} & \textbf{After correction: observed result} \\
\midrule
$S_1$ & 12 Elm Street; CURRENT & STALE direct root; $N_1$ (88 Oak Avenue) is the replacement \\
$D_1$: $S_1\to S_2$ & Gold: STRICT \texttt{DEPENDS\_ON} & Verified; invalidated $S_2$ at depth 1 \\
$S_2$ & Courier label from Elm Street; CURRENT & STALE \\
$D_2$: $S_2\to S_3$ & Gold: STRICT \texttt{DERIVED\_FROM} & Verified; invalidated $S_3$ at depth 2 \\
$S_3$ & Thursday estimate from label route; CURRENT & STALE \\
$D_3$: $S_3\to S_4$ & Gold: STRICT \texttt{AFFECTS\_ACTION} & Verified action dependency; invalidated the queued email at depth 3 \\
$S_4$ & Arrival email queued; CURRENT & STALE; original send gated for replanning \\
$S_5$ & Gift note included; CURRENT & CURRENT; unrelated to the failed address path \\
\bottomrule
\end{tabular}
\end{minipage}
\par\medskip

\begingroup
\raggedright
\noindent\textbf{Gold annotations:}
\texttt{gold\_\allowbreak direct\_\allowbreak invalidated\_\allowbreak states=[S1]},
\texttt{gold\_\allowbreak propagated\_\allowbreak invalidated\_\allowbreak states=[S2,S3,S4]},
\texttt{gold\_\allowbreak keep\_\allowbreak states=[S5]}, and
\texttt{gold\_\allowbreak action=REVISE\_\allowbreak PLAN}. Gold annotations specify the required outcomes; StateGraph selected and assessed independent support at runtime.

\noindent\textbf{Code basis:}
\texttt{stategraph/\allowbreak state/\allowbreak canonical\_\allowbreak state.py} (\texttt{CanonicalState});
\texttt{stategraph/\allowbreak state/\allowbreak streaming\_\allowbreak v3.py} (\texttt{revise}, \texttt{discover}, \texttt{supports}, \texttt{ingest});
\texttt{stategraph/\allowbreak state/\allowbreak revision\_\allowbreak v3.py} (\texttt{normalize\_\allowbreak revision\_\allowbreak output});
\texttt{stategraph/\allowbreak state/\allowbreak relation\_\allowbreak typing\_\allowbreak v3.py} (\texttt{\_\allowbreak valid\_\allowbreak type}, \texttt{normalize\_\allowbreak relation\_\allowbreak decision});
\texttt{stategraph/\allowbreak state/\allowbreak counterfactual\_\allowbreak verification\_\allowbreak v3.py} (\texttt{verification\_\allowbreak request}, \texttt{normalize\_\allowbreak verification});
\texttt{stategraph/\allowbreak state/\allowbreak alternative\_\allowbreak support\_\allowbreak v3.py} (\texttt{support\_\allowbreak request}, \texttt{normalize\_\allowbreak support});
\texttt{stategraph/\allowbreak state/\allowbreak cascade\_\allowbreak propagation\_\allowbreak v3.py} (\texttt{propagate});
\texttt{stategraph/\allowbreak state/\allowbreak premise\_\allowbreak gate\_\allowbreak v3.py} (\texttt{gate}).
Gold source: StateChangeBench annotations (50 cases),
SHA-256
\texttt{77d9c6c90e01a5bc\allowbreak 5c1838d64cf523bd\allowbreak 828c9a88c6f13711\allowbreak 586409b976f60725}.
\par
\endgroup

\end{document}